\documentclass[a4paper,fleqn]{cas-dc}
\usepackage[numbers,sort&compress]{natbib}
\usepackage{flushend}
\usepackage{amsmath}
\usepackage{amssymb}
\usepackage{booktabs}
\usepackage{multirow}
\usepackage{tabularx}
\RenewDocumentCommand\printorcid{}{}
\begin{document}
\let\WriteBookmarks\relax
\shorttitle{}
\shortauthors{S. Wang et al.}
\title[mode=title]{STEAM: Stable Self-Training with Elastic Matching and Adaptive Purification}
\author[1]{Shaoxiang Wang}
\author[1]{Kejia Zhang}
\cormark[1]
\ead{kejiazhang@hrbeu.edu.cn}
\author[1]{Haiwei Pan}
\author[1]{Jian Guan}
\author[2]{Lan Zhang}
\author[1]{Guozheng Zheng}

\affiliation[1]{organization={Harbin Engineering University, School of Computer Science and Technology},
    city={Harbin},
    country={China}}
\affiliation[2]{organization={Northeast Forestry University, School of Computer and Artificial Intelligence},
    city={Harbin},
    country={China}}
\cortext[cor1]{Corresponding author.}

\begin{abstract}
Cross-view geo-localization (CVGL) aims to perform visual localization by matching drone-view images with corresponding satellite images. Existing supervised methods rely on large-scale, manually annotated cross-view image pairs, which are costly and hard to scale. Many unsupervised approaches obtain pseudo-supervision through image-generation-based warm-up or clustering. These procedures may introduce distribution bias or propagate pseudo-label errors.
To address these limitations, we propose \textbf{STEAM}
(Stable Self-Training with Elastic Matching and Adaptive Purification),
a direct self-training framework for unsupervised cross-view
geo-localization.
It combines end-to-end representation learning with iterative pseudo-label refinement without image-generation-based warm-up or clustering. Specifically, the Stable Spatial-Aware Module constructs image descriptors through bounded nonlinear spatial weighting. Elastic Matching combines Bidirectional Top-\(K\) Soft Matching with Dynamic Threshold Filtering to accommodate many-to-one cross-view correspondences while filtering ambiguous candidates. Adaptive Purification maintains the pseudo-label repository through Confidence-Aware Updates, Age-Aware Updates, and Expired Label Removal. These operations respectively update assignments through confidence comparisons, track consecutive epochs without agreement or successful replacement, and discard entries whose age reaches a preset limit. On University-1652, STEAM achieves the highest R@1 and AP among the compared unsupervised methods in both retrieval directions. It also achieves competitive performance on SUES-200 and DenseUAV.
\end{abstract}
\maketitle

\section{Introduction}
Cross-view geo-localization (CVGL) aims to achieve GPS-free localization by matching a query image (e.g., a drone-view image) with its corresponding gallery image (e.g., a satellite-view image) in a large-scale image database~\cite{workman2015wide}. As a visual localization technique, CVGL has attracted increasing attention due to its broad applications in autonomous drone navigation, autonomous driving, disaster response, military reconnaissance, and smart cities~\cite{shetty2019uav}. However, substantial differences in viewpoint, resolution, and appearance between drone and satellite images make this task highly challenging~\cite{deuser2023sample4geo}.

Drone-satellite retrieval is particularly challenging because corresponding images need not depict the scene at comparable scales or with similar spatial coverage. Oblique drone views emphasize building facades and local structures, whereas satellite imagery emphasizes overhead layout. Nearby locations may share buildings, roads, or vegetation patterns, producing visually similar but geographically incorrect candidates. Moreover, multiple drone images can correspond to the same satellite location, so reliable correspondence discovery must accommodate many-to-one relationships~\cite{zheng2020university,zhu2023sues}.

Recent supervised methods have achieved remarkable progress by learning discriminative cross-view feature representations from manually annotated drone--satellite image pairs~\cite{durgam2024cross,zhang2023cross}. However, constructing large-scale paired datasets typically requires expensive sensing equipment and labor-intensive manual annotation, resulting in high data collection costs~\cite{li2024unleashing,li2025unsupervised}. This motivates unsupervised cross-view geo-localization (UCVGL). Unpaired drone and satellite images may be available even when their location-level associations have not been verified. Exploiting these collections requires discovering informative training pairs while avoiding ambiguous associations that can corrupt the learning process.

\begin{figure}[pos=tbp]
\centering 
\includegraphics[width=\columnwidth]{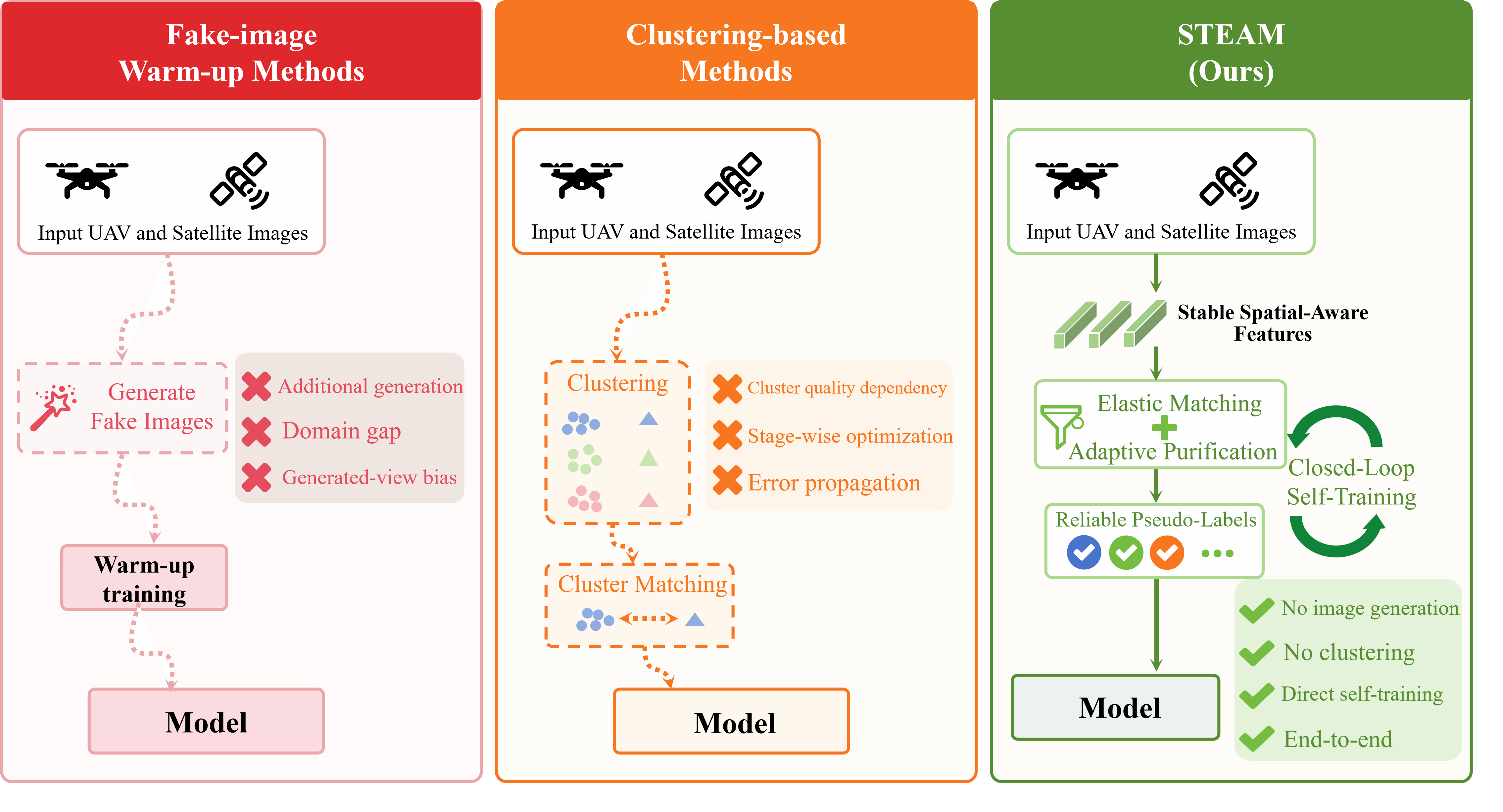}
\caption{Comparison of STEAM with representative UCVGL pipelines. STEAM combines end-to-end representation learning with iterative pseudo-label refinement, without image-generation-based warm-up or clustering.} 
\label{intro}
\end{figure}

Several UCVGL methods have recently been proposed, including fake-image warm-up methods~\cite{li2024unleashing} and clustering-based methods~\cite{wang2025coarse,chen2026uniabg,chen2026without}. Fake-image warm-up methods reduce viewpoint discrepancies by generating intermediate views, whereas clustering-based methods derive pseudo supervision from feature clusters and cross-view association. Despite promising results, the two approaches face different sources of unreliable supervision. As illustrated in Fig.~\ref{intro}, fake-image warm-up methods require an additional image-generation stage, and the generated views may introduce distribution bias into subsequent representation learning. In contrast, clustering-based methods depend on cluster quality to establish correspondences for many-to-one drone-to-satellite matching. Errors in early cluster assignments may propagate through subsequent training, causing noisy pseudo-labels to accumulate and degrade localization performance.

Other directions adapt frozen foundation models using unlabeled data~\cite{li2024learning} or align pretrained descriptors without training a task-specific network~\cite {lu2026vfm}. These approaches offer alternative ways to exploit pretrained knowledge, but controlling noisy supervision remains necessary when the encoder is updated through self-training. In this setting, correspondence estimation and feature learning are coupled: the training pairs depend on the representation they help shape.

These observations motivate direct self-training that discovers correspondences from the input drone and satellite images. Its effectiveness depends on three interacting requirements. First, image descriptors must preserve discriminative spatial information despite cross-view appearance changes. Second, correspondence discovery must balance pseudo-label coverage against reliability, since overly restrictive matching can discard valid many-to-one associations. Third, supervision must be revised as the representation evolves, because previously accepted pseudo-labels may become inconsistent with newly learned similarities. Addressing only one requirement leaves the remaining sources of error in the learning loop. A suitable framework should therefore coordinate representation learning, candidate discovery, and the maintenance of reliable supervision throughout training. This is especially relevant during early iterations, when confident yet incorrect matches can be repeatedly reused and constrain subsequent representation learning.

To address this gap, we propose STEAM (Stable Self-Training with Elastic Matching and Adaptive Purification), a self-training framework for UCVGL that combines end-to-end representation learning with iterative pseudo-label refinement. STEAM performs self-training directly on the input drone and satellite images without image-generation-based warm-up or clustering. First, the Stable Spatial-Aware Module (SSA) introduces bounded nonlinear spatial weighting to aggregate backbone features into descriptors for cross-view correspondence discovery. Second, Elastic Matching (ElMa) combines Bidirectional Top-$K$ Soft Matching and Dynamic Threshold Filtering to discover reliable many-to-one cross-view correspondences. Third, Adaptive Purification (AdPu) maintains a reliable pseudo-label repository through Confidence-Aware Update, Age-Aware Update, and Expired Label Removal. Together, these components form a closed-loop self-training framework that iteratively improves feature representations and pseudo-label quality.
The main contributions of this paper are summarized as follows:
\begin{itemize}
  \item  We propose STEAM, a framework for UCVGL that combines end-to-end representation learning with iterative pseudo-label refinement. It performs direct self-training without image-generation-based warm-up or clustering. We further introduce the Stable Spatial-Aware Module (SSA), which incorporates a bounded nonlinear mapping into multi-head spatial feature aggregation.

  \item We propose Elastic Matching (ElMa), a pseudo-label matching strategy that combines Bidirectional Top-$K$ Soft Matching with Dynamic Threshold Filtering to discover reliable many-to-one cross-view correspondences and improve pseudo-label coverage.

  \item  We design Adaptive Purification (AdPu) to maintain a reliable pseudo-label repository through Confidence-aware Update, Age-aware Update, and Expired Label Removal, thereby mitigating the accumulation of stale pseudo-labels during self-training.
\end{itemize}

\section{Related Work}

\subsection{Supervised Cross-View Geo-Localization}
Supervised CVGL learns geographically consistent representations from annotated cross-view image pairs. Early approaches use global descriptors, including CVM-Net~\cite{hu2018cvm}, which combines convolutional features with NetVLAD aggregation. VIGOR~\cite{zhu2021vigor} extends ground-to-aerial retrieval to unaligned query and reference locations, where multiple satellite images may cover the same query. University-1652~\cite{zheng2020university}, SUES-200~\cite{zhu2023sues}, and DenseUAV~\cite{dai2024vision} extend evaluation to drone imagery under different viewpoints, flight heights, and sampling densities. These benchmarks expose substantial changes in viewpoint, scale, and scene coverage that complicate descriptor learning even when geographic supervision is available.

Spatial alignment and contextual modeling help bridge the viewpoint gap. Orientation-aware learning~\cite{liu2019lending} incorporates directional information to reduce spatial ambiguity across heterogeneous viewpoints. SAFA~\cite{shi2019spatial} combines polar transformation with learned spatial embeddings, while Optimal Feature Transport~\cite{shi2020optimal} models local correspondences. View from Above~\cite{wang2024view} uses orthographic aerial geometry to aggregate features for ground-view pose estimation, including information from structures above the ground plane. LPN~\cite{wang2021each} preserves contextual information through square-ring partitioning, whereas FSRA~\cite{dai2021transformer} derives and aligns regions from transformer feature responses. View-synthesis approaches~\cite{regmi2019bridging,tian2021uav} reduce appearance differences through generated views, although synthesis alone does not imply an unpaired training protocol.

TransGeo~\cite{zhu2022transgeo} concentrates computation on informative patches through attention-guided cropping, while GeoDTR~\cite{zhang2023crossview} disentangles geometric layout from appearance. Sample4Geo~\cite{deuser2023sample4geo} emphasizes symmetric contrastive learning and hard negative sampling. DAC~\cite{xia2024enhancing} and CAMP~\cite{wu2024camp} further explore domain alignment, scene consistency, and position-aware features. ConGeo~\cite{mi2024congeo} also addresses variations in ground-view observations to improve representation robustness. These methods improve descriptor discrimination using annotated correspondences; STEAM instead learns from iteratively discovered pseudo-labels.

\subsection{Unsupervised Cross-View Geo-Localization}
Unsupervised CVGL must jointly improve representations and estimate correspondences without annotated cross-view pairs. Incorrect initial matches can reinforce misleading feature similarities, making pseudo-label selection and subsequent refinement central to stable training. The risk of reinforcing erroneous pseudo-labels is also documented as confirmation bias in semi-supervised image classification~\cite{arazo2020pseudolabeling}. Li et al.~\cite{li2024unleashing} address initialization through correspondence-free projection and image-generation-based warm-up. Their framework transforms ground panoramas into bird's-eye views and generates satellite-style images using CycleGAN before learning from real cross-view collections.

Clustering provides another source of supervision, drawing on cluster-level contrastive learning~\cite{caron2020unsupervised,dai2022cluster}. C2F~\cite{wang2025coarse} combines intra-view clustering with cross-view location alignment and intermediate-state transition. UniABG~\cite{chen2026uniabg} couples adversarial view bridging with heterogeneous graph filtering. DMNIL~\cite{chen2026without} uses clustering-based pseudo-labels, dynamic hierarchical memory, and neighborhood-driven information consistency learning. Although these strategies strengthen feature discrimination and association, clustering quality remains important to the supervision they produce.

PPLR~\cite{shao2026pplr} studies pseudo-label control for ground--satellite retrieval through reciprocal neighborhood expansion, margin-based filtering, and consistency-guided refinement after cold-start initialization. It highlights the need to balance candidate coverage and label reliability. STEAM addresses these concerns for many-to-one drone--satellite matching using Elastic Matching and Adaptive Purification, without clustering or image-generation-based warm-up. Its repository maintains and updates reliable correspondences throughout self-training.

\subsection{Foundation Models and Training-Free Geo-Localization}
Foundation models offer alternatives to learning encoders from paired geographic data. AnyLoc~\cite{keetha2023anyloc} combines pretrained self-supervised features with unsupervised aggregation for visual place recognition. EM-CVGL~\cite{li2024learning} instead trains an adapter over a frozen foundation model using expectation-maximization-based pseudo-labeling and information consistency. It removes paired supervision but retains task-specific parameter learning.

VFM-Loc~\cite{lu2026vfm} performs zero-shot matching through hierarchical feature aggregation and statistical manifold alignment. Li et al.~\cite{li2025unsupervised} reconstruct scenes from multiple UAV images using 3D Gaussian Splatting and render virtual satellite views. Their approach avoids paired retrieval-network training but requires scene reconstruction and optimization, differing from training-free descriptor alignment and learned single-image retrieval. Consequently, comparisons should distinguish image-level retrieval from localization requiring multiple observations of the same scene.

Multimodal approaches broaden the available supervision. CrossText2Loc~\cite{ye2025where} retrieves satellite imagery or map representations from natural-language descriptions. GeoBridge~\cite{song2026geobridge} aligns drone, street-view, and satellite features using textual semantic anchors and geographically aligned multi-view data. These settings differ from STEAM's focus on end-to-end encoder optimization and reliable correspondence discovery from unpaired drone--satellite images.

\section{Method}

We propose STEAM (Stable Self-Training with Elastic Matching and Adaptive Purification), a self-training framework for UCVGL that combines end-to-end representation learning with iterative pseudo-label refinement. Given unpaired drone and satellite images, STEAM iteratively generates and refines pseudo-labels to align the cross-view feature space. We first formulate the problem and present the overall framework, followed by the Stable Spatial-Aware Module, Elastic Matching, Adaptive Purification, and the optimization objective.

\begin{figure*}[width=\textwidth,pos=t]
  \centering
\includegraphics[width=0.95\textwidth]{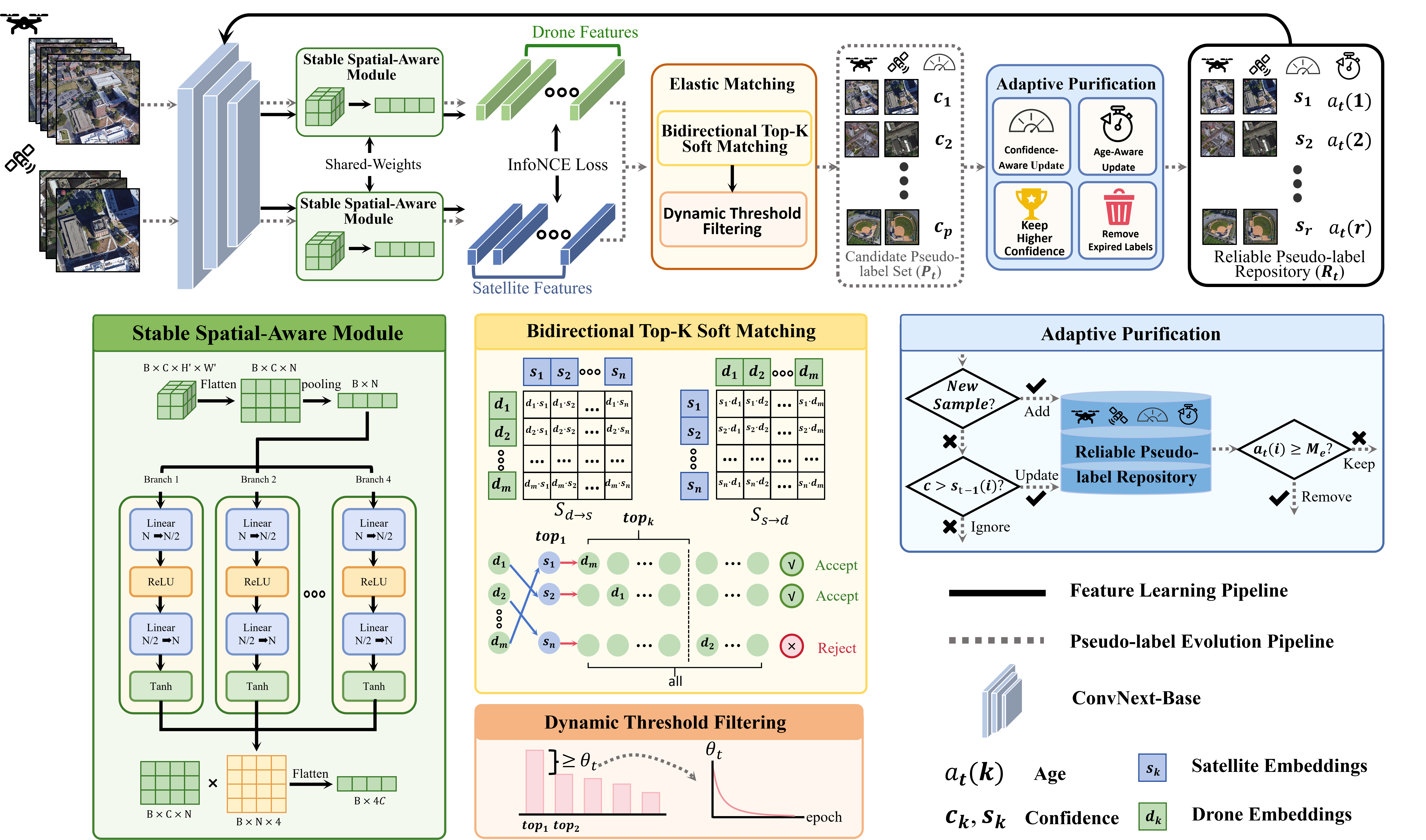}
\caption{Overview of STEAM. The top row shows the closed-loop self-training pipeline, and the bottom row details its three components. A shared backbone and SSA extract cross-view representations $Z_t^d$ and $Z_t^s$. ElMa constructs the candidate pseudo-label set $P_t$ using Bidirectional Top-$K$ Soft Matching and Dynamic Threshold Filtering. AdPu updates the reliable pseudo-label repository $R_t$ through Confidence-aware Update, Age-aware Update, and Expired Label Removal. Training pairs constructed from $R_t$ are used to optimize the model with the symmetric InfoNCE loss.}
\label{model}
\end{figure*}

\textbf{Problem Formulation.} We consider an unlabeled cross-view geo-localization dataset consisting of drone images $X_d=\{x_i^d\}_{i=1}^{N_d}$ and satellite images $X_s=\{x_j^s\}_{j=1}^{N_s}$, where $N_d$ and $N_s$ denote the numbers of drone and satellite images, respectively. Ground-truth cross-view correspondences are not inputs to pseudo-label mining, mini-batch construction, or gradient-based optimization. The objective of UCVGL is to learn discriminative cross-view feature representations without cross-view correspondence annotations for cross-view image retrieval. Since multiple drone images may correspond to the same satellite image, UCVGL is naturally formulated as a many-to-one cross-view matching problem.

\subsection{Overview of STEAM}

As discussed in the Introduction, many UCVGL methods rely on fake-image generation or clustering initialization because direct self-training suffers from noisy pseudo-label supervision. To address this issue, STEAM combines end-to-end representation learning with iterative pseudo-label refinement through three interacting components (Fig.~\ref{model}). The Stable Spatial-Aware Module (SSA) constructs image descriptors through bounded nonlinear spatial weighting, Elastic Matching (ElMa) discovers reliable many-to-one correspondences, and Adaptive Purification (AdPu) progressively maintains a reliable pseudo-label repository. Together, these components form a closed-loop self-training framework that iteratively improves feature representations and pseudo-label quality.

At iteration $t$, the SSA first extracts feature representations from the drone and satellite images:
\begin{equation}
(Z_t^d,Z_t^s)=f_{\mathrm{SSA}}(X_d,X_s),
\end{equation}
ElMa then applies Bidirectional Top-$K$ Soft Matching and Dynamic Threshold Filtering to generate a candidate pseudo-label set:
\begin{equation}
P_t=f_{\mathrm{ElMa}}(Z_t^d,Z_t^s),
\end{equation}

where $P_t$ denotes the candidate pseudo-label set. AdPu then updates the reliable pseudo-label repository using the candidate pseudo-labels and the repository from the previous iteration:
\begin{equation}
R_t=f_{\mathrm{AdPu}}(P_t,R_{t-1}),
\end{equation}
where $R_t$ denotes the reliable pseudo-label repository. Training pairs are constructed from $R_t$, and the model is optimized using the symmetric InfoNCE loss.

STEAM jointly trains the shared backbone and SSA end to end using the symmetric InfoNCE objective. ElMa and AdPu update discrete pseudo-label assignments between optimization phases, with the assignments fixed within each phase.

\subsection{Stable Spatial-Aware Module}
Direct self-training constructs cross-view pseudo-correspondences
from the current image descriptors without image-generation-based
warm-up or clustering.
Because these correspondences subsequently supervise the feature
extractor, errors in correspondence discovery can propagate
through the training loop.
This motivates controlling the spatial weighting used to
construct the descriptors.
Building on spatial-aware feature aggregation~\cite{shi2019spatial},
SSA retains the multi-head aggregation structure and introduces
a bounded nonlinear mapping for spatial weight generation.
ReLU introduces nonlinearity between the two linear projections,
while tanh restricts each output weight to $(-1,1)$.
Relative to the corresponding two-layer linear mapping with
the same dimensions, these activation functions introduce
no additional trainable parameters.
The resulting descriptors are used by ElMa to discover
cross-view pseudo-correspondences.

Given an input image $x\in\mathbb{R}^{H\times W\times3}$, a ConvNeXt-Base backbone~\cite{liu2022convnext} is first employed to extract dense feature maps:
\begin{equation}
x_{b}=f_{\mathrm{backbone}}(x)\in\mathbb{R}^{C\times H'\times W'},
\end{equation}
where $C$ denotes the number of channels, and $H'$ and $W'$ are the spatial dimensions. The feature map is flattened to $x_{f}=f_{\mathrm{Flatten}}(x_{b})\in\mathbb{R}^{C\times N}$ with $N=H'\times W'$. Max pooling along the channel dimension then produces a spatial descriptor:
\begin{equation}
x_p=f_{\mathrm{pooling}}(x_{f})\in\mathbb{R}^{N}.
\end{equation}

$M$ separately parameterized attention heads are used to generate spatial attention responses, where $M=4$ in all experiments. For the $m$-th attention head
($m=1,\ldots,M$),
\begin{equation}
a_m=\tanh\left(W_2^m\cdot\mathrm{ReLU}\left(W_1^m x_p\right)\right)\in\mathbb{R}^{N},
\end{equation}
where $W_1^m\in\mathbb{R}^{\frac{N}{2}\times N}$ and
$W_2^m\in\mathbb{R}^{N\times\frac{N}{2}}$ are learnable weight matrices; the bias terms are omitted for clarity. Here, ReLU introduces nonlinearity between the two linear projections, while tanh restricts each element of $a_m$ to $(-1,1)$. The attention responses are stacked column-wise to form $A=[a_1,a_2,\ldots,a_M]\in\mathbb{R}^{N\times M},$

The attention matrix is then used to aggregate spatial features:
\begin{equation}
x_h=x_fA\in\mathbb{R}^{C\times M}.
\end{equation}

The aggregated feature matrix is transposed, flattened, and L2-normalized to obtain the final representation $z\in\mathbb{R}^{CM}$. The ConvNeXt-Base backbone and SSA are shared across views, producing feature matrices
 $Z_d=\{z_i^d\}, Z_s=\{z_i^s\}$.

\subsection{Elastic Matching}
The inherent many-to-one relationship between drone and satellite images makes strict mutual nearest-neighbor matching overly restrictive, because multiple drone images may correspond to the same satellite image. To address this issue, we propose Elastic Matching (ElMa), which combines Bidirectional Top-$K$ Soft Matching and Dynamic Threshold Filtering. Bidirectional Top-$K$ Soft Matching relaxes the reverse matching constraint to improve pseudo-label coverage, while Dynamic Threshold Filtering removes ambiguous candidate correspondences according to a dynamically adjusted similarity margin.

\textbf{Bidirectional Top-$K$ Soft Matching.}
Given feature representations $Z_d$ and $Z_s$, similarity matrices are computed as
 $S_{d\rightarrow s} = Z_d Z_s^\top,\quad S_{s\rightarrow d} = Z_s Z_d^\top$. For a drone image $i$, its nearest satellite image is determined by $j_1 = \arg\max_j S_{d\rightarrow s}(i,j)$. To avoid enforcing strict mutual nearest-neighbor constraints, we only require drone $i$ to appear in the Top-$K$ most similar drones of satellite $j_1$:

\begin{equation}
\mathrm{Match}(i) =
\mathbb{I}\big[i \in \mathrm{Top}\text{-}K (S_{s\rightarrow d}(j_1,:))\big],
\end{equation}
where $K$ denotes the number of nearest neighbors. The influence of $K$ is further analyzed in Fig.~\ref{k_analysis}.

The reverse Top-K condition permits many-to-one correspondences while retaining a hard top-1 satellite assignment for each drone image.

\textbf{Dynamic Threshold Filtering.}
Although Bidirectional Top-$K$ Soft Matching improves pseudo-label coverage, incorrect matches may still exist. To further eliminate low-confidence samples, we adopt a dynamic threshold filtering strategy~\cite{li2024unleashing}. The similarity margin of drone $i$ is
\begin{equation}
\delta_i = S_{d\rightarrow s}(i,j_1) - S_{d\rightarrow s}(i,j_2),
\end{equation}
where $j_1$ and $j_2$ denote the indices of the most similar and second most similar satellite images, respectively. A pseudo-label is accepted only when

\begin{equation}
\mathrm{Confident}(i) = \mathbb{I}[\delta_i \ge \theta_t],
\end{equation}
where the dynamic threshold follows a cosine decay schedule:

\begin{equation}
\theta_t = \frac{\theta_0}{2}\left(1 + \cos\left(\frac{\pi(t-1)}{T - 1}\right)\right), \quad \theta_0 = 0.05,
\end{equation}
where $t=1,\ldots,T$ denotes the current training epoch and $T=20$ is the total number of training epochs. 

\textbf{Candidate Pseudo-label Construction.}
The drone images satisfying both the matching and confidence constraints are selected to form the candidate pseudo-label set:

\begin{equation}
P_t = \{(i,j_1,c_i)\mid \mathrm{Match}(i) \land \mathrm{Confident}(i)\},
\end{equation}
where $c_i = S_{d\rightarrow s}(i,j_1)$ denotes the confidence score associated with the pseudo-label.

The two tests serve complementary purposes. Reverse-rank membership checks whether a drone remains a plausible candidate from the satellite perspective, while the margin measures ambiguity between satellite alternatives. Their conjunction accommodates multiple drone views while retaining a query-specific ambiguity filter.

\subsection{Adaptive Purification}

Pseudo-labels may become outdated as feature representations evolve during iterative self-training. To maintain a reliable pseudo-label repository, we propose Adaptive Purification (AdPu), which updates historical pseudo-labels through Confidence-aware Update, Age-aware Update, and Expired Label Removal.

At iteration $t$, Elastic Matching generates a candidate pseudo-label set $P_t=\{(i,j,c)\}$, where $i$, $j$, and $c$ denote the drone image index, matched satellite image index, and confidence score, respectively. AdPu maintains a repository $R_t=\{(i,y_t(i),s_t(i),a_t(i))\}$. Here, $y_t(i)$ is the pseudo-label of drone image $i$, represented by the satellite image index assigned through pseudo-label matching and repository updates. The values $s_t(i)$ and $a_t(i)$ denote the confidence score and age of this pseudo-label, respectively. Specifically, $a_t(i)$ counts the consecutive epochs in which
the stored pseudo-label is neither supported by an agreeing candidate in $P_t$ nor replaced by the confidence-aware update.
The repository is initialized as $R_0=\varnothing$.

\textbf{Confidence-Aware Update.}
For each candidate pair $(i,j,c)\in P_t$, we compare its confidence $c$ with the confidence stored in $R_{t-1}$. If $c>s_{t-1}(i)$ or drone image $i$ does not exist in $R_{t-1}$, the entry is updated to $(j,c)$; otherwise, the existing entry is retained:

{\small
\begin{equation}
(y_t(i),s_t(i)) =
\begin{cases}
(j,c), & c>s_{t-1}(i)\ \text{or}\ i\notin R_{t-1},\\
(y_{t-1}(i),s_{t-1}(i)), & \text{otherwise}.
\end{cases}
\end{equation}
}

\textbf{Age-Aware Update.}
After the confidence-aware update, let $y_t(i)$ denote the
current stored pseudo-label of drone image $i$, before expiration
is applied. The age is updated as
\begin{equation}
a_t(i)=
\begin{cases}
0,
& \exists\,(i,j,c)\in P_t:\ j=y_t(i),\\
a_{t-1}(i)+1,
& \text{otherwise}.
\end{cases}
\end{equation}
Newly inserted entries satisfy the first condition and are
initialized with age zero; the second case applies to
previously stored entries.
An agreeing candidate or a successful replacement resets
the age, whereas a conflicting candidate that does not
replace the stored pseudo-label increases its age.
The age also increases when no candidate is available.

\textbf{Expired Label Removal.}
Any entry satisfying
$
a_t(i) \ge M_e
$
is considered expired and removed from \(R_t\).
After updating and removing invalid entries, the repository is obtained as:
\begin{equation}
R_t = \mathrm{AdPu}(P_t, R_{t-1}).
\end{equation}

Together, these updates remove an unreplaced historical pseudo-label
after $M_e$ consecutive epochs without an agreeing candidate.
Repeated conflicting candidates that fail to replace the
stored pseudo-label therefore cannot indefinitely reset its age.

ElMa evaluates current candidate correspondences, while AdPu maintains their validity across iterations. An agreeing candidate can reset the age even when its confidence is below the stored score. A lower-confidence conflicting candidate neither replaces the label nor resets its age. Thus, confidence controls replacement, and age records how long the retained assignment has lacked current support.

\subsection{Optimization Objective}

From the pseudo-label repository $R_t$, we construct a mini-batch of $B$ cross-view pairs $\{(x_i^d, x_{y_i}^s)\}_{i=1}^{B}$.
Here, $y_i$ is the pseudo-label stored in $R_t$ for the sampled drone image, specifying its predicted satellite match.
Drone images with the same pseudo-label value are not sampled together.
This sampling rule uses only pseudo-labels in $R_t$ and requires no ground-truth correspondences. The normalized feature matrices $Z_d, Z_s \in \mathbb{R}^{B\times d}$ are used to compute the cross-view similarity $S = Z_d Z_s^\top / \tau$. The temperature is parameterized as $\tau=\exp(-\gamma)$,
where $\gamma$ is a learnable scalar initialized to $\log(1/0.07)$.
It is optimized jointly with the backbone and SSA.
We use the pseudo-label assignments as hard targets without label smoothing.
The model is optimized with a symmetric form of the InfoNCE objective~\cite{oord2018cpc}:

{\small
\begin{equation}
\mathcal{L} = -\frac{1}{2B}\sum_{i=1}^{B}
\left[
\log \frac{\exp(S_{ii})}{\sum_{j=1}^{B}\exp(S_{ij})}
+
\log \frac{\exp(S_{ii})}{\sum_{j=1}^{B}\exp(S_{ji})}
\right].
\end{equation}
}

The distinct-pseudo-label sampling rule is consistent with the diagonal targets in this symmetric objective. Including two pairs with the same satellite assignment would introduce additional matching entries that the loss treats as negatives. Distinct assignments avoid this conflict within a batch while allowing many-to-one correspondences across batches.

By optimizing this objective with pseudo-labels from Adaptive Purification, STEAM progressively improves cross-view feature alignment within a unified closed-loop self-training framework.

\subsection{Training Procedure}

Training alternates between pseudo-label refinement and representation
optimization once per epoch, for a total of $T=20$ epochs.
The shared backbone is initialized with ImageNet-pretrained weights,
and the linear layers in SSA use the default PyTorch initialization.
The pseudo-label repository is initialized as $R_0=\varnothing$.

At the beginning of epoch $t$, the model obtained from the preceding
epoch extracts descriptors for all drone and satellite images.
Feature extraction uses evaluation mode, deterministic preprocessing,
and no gradient computation.
ElMa constructs the candidate set $P_t$, and AdPu updates the
repository from $R_{t-1}$ to $R_t$.
In the first epoch, mining uses the initialized model before any
contrastive optimization, and the resulting candidates populate
the initially empty repository.

The pseudo-label assignments in $R_t$ remain fixed during the subsequent
optimization phase.
Mini-batches are formed from $R_t$, and their descriptors are
recomputed with training augmentations for each forward pass.
The model undergoes one optimization epoch using the symmetric
InfoNCE objective before the next full-dataset mining step.
Each mini-batch contains at most 80 pairs with distinct pseudo-label
values from $R_t$; smaller batches are used when fewer distinct
pseudo-label values are available.
Only batches containing at least two pairs are optimized.
If an updated repository contains fewer than two distinct pseudo-label
values, the run is terminated and recorded as an
insufficient-supervision failure.

\begin{table*}[width=\FullWidth,pos=t]
\centering
\caption{Performance comparison on University-1652. Best results within each learning category are marked in \textbf{bold}.}
\label{tab:sota_u1652}
\begin{tabularx}{\textwidth}{c c >{\centering\arraybackslash}X >{\centering\arraybackslash}X >{\centering\arraybackslash}X >{\centering\arraybackslash}X}
\toprule
\multirow{2}{*}{} & \multirow{2}{*}[-0.5ex]{Method} & \multicolumn{2}{c}{Drone $\rightarrow$ Satellite} & \multicolumn{2}{c}{Satellite $\rightarrow$ Drone} \\ \cmidrule(r){3-4} \cmidrule(r){5-6}
& & R@1 & AP & R@1 & AP \\
\midrule
\multirow{8}{*}{Supervised} & Zheng et al.~\cite{zheng2020university} & 59.69 & 64.80 & 73.18 & 59.40 \\
 & PCL \cite{tian2021uav} & 79.47 & 83.63 & 87.69 & 78.51 \\
 & FSRA~\cite{dai2021transformer} & 82.25 & 84.82 & 87.87 & 81.53 \\
 & MCCG~\cite{shen2023mccg} & 89.28 & 91.01 & 94.29 & 89.29 \\
 & Sample4Geo~\cite{deuser2023sample4geo} & 92.65 & 93.81 & 95.14 & 91.39 \\
 & DAC~\cite{xia2024enhancing} & 94.67 & 95.50 & 96.43 & 93.79 \\
 & QDFL~\cite{hu2025query} & 95.00 & 95.83 & \textbf{97.15} & 94.57 \\
 & AGEN~\cite{xu2025agen} & \textbf{95.43} & \textbf{96.18} & 96.72 & \textbf{95.52} \\
\midrule
\multirow{5}{*}{Unsupervised} & Li et al.~\cite{li2024unleashing} & 70.29 & 74.93 & 79.03 & 61.03 \\
 & Wang et al.~\cite{wang2025coarse} & 85.95 & 90.33 & 94.01 & 82.66 \\
 & DMNIL~\cite{chen2026without} & 90.17 & 91.67 & 95.01 & 88.95 \\
 & UniABG~\cite{chen2026uniabg} & 93.62 & 94.61 & 95.43 & 93.29 \\
 & STEAM & \textbf{95.26} & \textbf{96.08} & \textbf{96.01} & \textbf{94.21} \\
\bottomrule
\end{tabularx}
\end{table*}

\section{Experiments}

\subsection{Datasets and Experimental Settings}

\textbf{Datasets.} We evaluate STEAM on three benchmarks: University-1652~\cite{zheng2020university}, SUES-200~\cite{zhu2023sues}, and DenseUAV~\cite{dai2024vision}. University-1652 contains 1,652 buildings across 72 universities, with 701/951 buildings for training/testing. Each location has 54 synthetic drone-view images and one satellite image. SUES-200 comprises 200 scenes with drone images captured at four altitudes (150m, 200m, 250m, 300m), each containing 10,000 drone images and 200 satellite images; 60\% of the data is used for training. DenseUAV provides densely sampled low-altitude drone imagery at 80, 90, and 100 m from 14 university campuses, together with multi-temporal satellite imagery. Following standard protocols, we report Drone-to-Satellite (D2S) and Satellite-to-Drone (S2D) retrieval using Recall@1 (R@1) and Average Precision (AP) on University-1652 and SUES-200. For DenseUAV, we report D2S R@1 and Recall@5 (R@5) at each height.

\textbf{Implementation Details.}
All experiments are implemented in PyTorch on a single NVIDIA A100 GPU.
We use an ImageNet-pretrained ConvNeXt-Base backbone and input images
of size $384\times384$.
ImageNet normalization uses mean $(0.485,0.456,0.406)$ and standard
deviation $(0.229,0.224,0.225)$, following previous
works~\cite{chen2026uniabg,li2024unleashing,deuser2023sample4geo,xia2024enhancing}.
The model is trained for 20 epochs using AdamW with an initial
learning rate of $1\times10^{-4}$, $(\beta_1,\beta_2)=(0.9,0.999)$,
and $\epsilon=10^{-8}$.
Weight decay is set to $0.01$ for convolutional and linear weights;
biases, normalization parameters, and the temperature parameter
$\gamma$ are excluded from weight decay.
The cosine learning-rate scheduler is stepped once at the end of
each epoch, with a minimum learning rate of zero and no
learning-rate warm-up.
The maximum batch size is 80 cross-view pairs, and automatic
mixed precision is used during training.
The learnable temperature is initialized to $0.07$.
For Elastic Matching, we set $K=50$ and $\theta_0=0.05$,
with the threshold decaying over the same 20 epochs.
For Adaptive Purification, the maximum pseudo-label age is $M_e=1$.
For the University-1652 analyses, 10\% of the original training
buildings are held out for validation.
The D2S Recall@1 in Figs.~\ref{k_analysis} and~\ref{fig:age_analysis}
is evaluated on this validation set and guides hyperparameter
and checkpoint selection.
Pseudo-label quantity and accuracy in
Figs.~\ref{fig:ablation_pseudo}--\ref{fig:age_analysis}
are computed over the combined training and validation sets.
The main University-1652 and SUES-200 results use random seed 42. Additional University-1652 runs with seeds 43 and 2026 are reported in Table~\ref{tab:random_seeds}.

\textbf{Data Augmentation.}
During optimization, each view independently undergoes random
resized cropping to $384\times384$.
The crop area fraction is sampled from $[0.8,1.0]$, and the
aspect ratio is sampled from $[0.9,1.1]$ using RandomResizedCrop.
Bicubic interpolation with antialiasing is used for resizing.
Horizontal flipping is applied with probability $0.5$, using
the same flip decision for both images in a training pair.
These are the only stochastic training augmentations, and
ImageNet normalization is applied afterward.
For pseudo-label mining and retrieval evaluation, images are
deterministically resized to $384\times384$ and normalized,
without random cropping or flipping.

Before training on DenseUAV, we generate 50 augmented instances per original drone training image using the augmentation operations described above. The resulting training pool is fixed throughout training, and each instance is treated as a separate unlabeled sample. The common source-image identities of these instances are not used for pseudo-label mining, mini-batch construction, or model optimization. Pseudo-label mining applies deterministic preprocessing to this fixed augmented pool.
\subsection{Comparison with State-of-the-Art Methods}
We compare STEAM with existing supervised and unsupervised methods on University-1652, SUES-200, and DenseUAV. Tables~\ref{tab:sota_u1652},~\ref{tab:sota_sues200}, and~\ref{tab:sota_denseuav} report the results under the methods' respective backbone configurations.

\begin{table*}[width=\FullWidth,pos=t]
\centering
\caption{Performance comparison on SUES-200. Best results within each learning category are marked in \textbf{bold}. $^*$ denotes unsupervised pretraining on University-1652 followed by unsupervised fine-tuning on SUES-200.}
\label{tab:sota_sues200}
{\small
\setlength{\tabcolsep}{4pt}
\begin{tabularx}{\textwidth}{c c >{\centering\arraybackslash}X>{\centering\arraybackslash}X>{\centering\arraybackslash}X>{\centering\arraybackslash}X>{\centering\arraybackslash}X>{\centering\arraybackslash}X>{\centering\arraybackslash}X>{\centering\arraybackslash}X}
\toprule
\multicolumn{10}{c}{\textbf{Drone $\rightarrow$ Satellite}}\\ \midrule
& \multirow{2}{*}[-0.5ex]{Method} & \multicolumn{2}{c}{150m} & \multicolumn{2}{c}{200m} & \multicolumn{2}{c}{250m} & \multicolumn{2}{c}{300m}\\
\cmidrule(r){3-4} \cmidrule(r){5-6} \cmidrule(r){7-8} \cmidrule(r){9-10}
& & R@1 & AP & R@1 & AP & R@1 & AP & R@1 & AP\\
\midrule
\multirow{6}{*}{Supervised} & SUES-200~\cite{zhu2023sues} & 55.65 & 61.92 & 66.78 & 71.55 & 72.00 & 76.43 & 74.05 & 78.26 \\
 & FSRA~\cite{dai2021transformer} & 68.25 & 73.45 & 83.00 & 85.99 & 90.68 & 92.27 & 91.95 & 93.46 \\
 & MCCG~\cite{shen2023mccg} & 82.22 & 85.47 & 89.38 & 91.41 & 93.82 & 95.04 & 95.07 & 96.20 \\
 & DAC~\cite{xia2024enhancing} & \textbf{96.80} & \textbf{97.54} & 97.48 & 97.97 & 98.20 & 98.62 & 97.58 & 98.14 \\
 & AGEN~\cite{xu2025agen} & 94.38 & 95.58 & 91.78 & 93.35 & 95.75 & 96.65 & 97.12 & 97.81 \\
 & QDFL~\cite{hu2025query} & 93.97 & 95.42 & \textbf{98.25} & \textbf{98.67} & \textbf{99.30} & \textbf{99.48} & \textbf{99.31} & \textbf{99.48} \\
\midrule
\multirow{6}{*}{Unsupervised} & Wang et al.~\cite{wang2025coarse} & 76.90 & 84.95 & 87.88 & 92.60 & 92.98 & 95.66 & 95.10 & 96.92 \\
 & UniABG~\cite{chen2026uniabg} & 92.40 & 93.95 & 97.32 & 97.92 & 98.07 & 98.55 & 98.67 & 98.98 \\
 & DMNIL~\cite{chen2026without} & 87.00 & 89.20 & 93.98 & 95.11 & 96.50 & 97.23 & 97.23 & 97.86 \\
 & DMNIL$^*$~\cite{chen2026without} & 90.98 & 92.66 & 95.03 & 96.14 & 97.45 & 98.03 & 98.23 & 98.64 \\
 & STEAM & 94.45 & 95.56 & 97.43 & 97.88 & 98.63 & 98.96 & 99.10 & 99.33 \\
 & STEAM$^*$ & \textbf{97.43} & \textbf{97.80} & \textbf{98.10} & \textbf{98.44} & \textbf{99.13} & \textbf{99.29} & \textbf{99.83} & \textbf{99.87} \\
\midrule
\addlinespace[2pt]
\multicolumn{10}{c}{\textbf{Satellite $\rightarrow$ Drone}}\\ \midrule
& \multirow{2}{*}[-0.5ex]{Method} & \multicolumn{2}{c}{150m} & \multicolumn{2}{c}{200m} & \multicolumn{2}{c}{250m} & \multicolumn{2}{c}{300m}\\
\cmidrule(r){3-4} \cmidrule(r){5-6} \cmidrule(r){7-8} \cmidrule(r){9-10}
& & R@1 & AP & R@1 & AP & R@1 & AP & R@1 & AP\\
\midrule
\multirow{6}{*}{Supervised} & SUES-200~\cite{zhu2023sues} & 75.00 & 55.46 & 85.00 & 66.05 & 86.25 & 69.94 & 88.75 & 74.46 \\
 & FSRA~\cite{dai2021transformer} & 83.75 & 76.67 & 90.00 & 85.34 & 93.75 & 90.17 & 95.00 & 92.03 \\
 & MCCG~\cite{shen2023mccg} & 97.50 & 93.63 & \textbf{98.75} & 96.70 & 98.75 & 98.28 & 98.75 & 98.05 \\
 & DAC~\cite{xia2024enhancing} & 97.50 & 94.06 & \textbf{98.75} & 96.66 & 98.75 & 98.09 & 98.75 & 97.87 \\
 & AGEN~\cite{xu2025agen} & 97.50 & 92.58 & 97.50 & 93.11 & 96.25 & 94.40 & 96.52 & 95.36 \\
 & QDFL~\cite{hu2025query} & \textbf{98.75} & \textbf{95.10} & \textbf{98.75} & \textbf{97.92} & \textbf{100.00} & \textbf{99.07} & \textbf{100.00} & \textbf{99.07} \\
\midrule
\multirow{6}{*}{Unsupervised} & Wang et al.~\cite{wang2025coarse} & 87.50 & 74.81 & 92.50 & 87.15 & 96.25 & 91.20 & 98.75 & 94.52 \\
 & UniABG~\cite{chen2026uniabg} & \textbf{98.75} & 91.54 & 98.75 & 97.06 & \textbf{100.00} & \textbf{98.32} & 98.75 & 97.58 \\
 & DMNIL~\cite{chen2026without} & 95.00 & 85.46 & 97.50 & 92.72 & \textbf{100.00} & 95.43 & \textbf{100.00} & 96.39 \\
 & DMNIL$^*$~\cite{chen2026without} & 96.25 & 90.05 & 97.50 & 94.80 & 98.75 & 96.82 & 98.75 & 97.33 \\
 & STEAM & 97.50 & 93.16 & 97.50 & 95.01 & \textbf{100.00} & 97.73 & 97.50 & 97.66 \\
 & STEAM$^*$ & \textbf{98.75} & \textbf{95.61} & \textbf{100.00} & \textbf{97.77} & 98.75 & 98.12 & \textbf{100.00} & \textbf{99.32} \\
\bottomrule
\end{tabularx}
}
\end{table*}

\textbf{University-1652.} STEAM achieves 95.26\%/96.08\% R@1/AP for D2S and 96.01\%/94.21\% for S2D, surpassing all compared unsupervised methods in Table~\ref{tab:sota_u1652}. Relative to UniABG, D2S R@1 and AP increase by 1.64 and 1.47 percentage points, respectively. The corresponding S2D gains are 0.58 and 0.92 percentage points. The improvements therefore extend to both retrieval directions and both metrics, with a larger R@1 gain for D2S. STEAM also approaches the best supervised D2S results, falling below AGEN by 0.17 percentage points in R@1 and 0.10 in AP. These comparisons use the reported model configurations, while Table~\ref{tab:ablation} examines individual design changes within STEAM.

\textbf{SUES-200.} Table~\ref{tab:sota_sues200} evaluates retrieval across four flight heights. Without University-1652 pretraining, STEAM's D2S R@1 increases from 94.45\% at 150 m to 99.10\% at 300 m. Its S2D R@1 reaches 100\% at 250 m, although the S2D results do not increase monotonically with height. Unsupervised pretraining followed by target-dataset fine-tuning further improves all four D2S R@1 scores. The largest gain is 2.98 percentage points at 150 m, compared with 0.67, 0.50, and 0.73 at the other heights. STEAM$^*$ reaches 99.83\% D2S R@1 at 300 m, exceeding the best supervised result of 99.31\%. It also achieves 100\% S2D R@1 at 200 and 300 m. The benefit is not uniform: at 250 m, S2D R@1 decreases from 100\% to 98.75\%, while AP increases from 97.73\% to 98.12\%. These results characterize transfer with unsupervised fine-tuning on SUES-200.

\begin{table*}[width=\FullWidth,pos=t]
\centering
\caption{Drone-to-Satellite retrieval performance on DenseUAV (\%). Best reported results within each learning category are marked in \textbf{bold}. STEAM is trained and evaluated separately at each height. For STEAM, each original drone training image is expanded into 50 augmented instances, which are treated as separate unlabeled samples.}
\label{tab:sota_denseuav}
{\small
\setlength{\tabcolsep}{2pt}
\begin{tabularx}{\textwidth}{c c *{6}{>{\centering\arraybackslash}X}}
\toprule
\multirow{2}{*}{} & \multirow{2}{*}{Method} & \multicolumn{2}{c}{80 m} & \multicolumn{2}{c}{90 m} & \multicolumn{2}{c}{100 m} \\
\cmidrule(lr){3-4} \cmidrule(lr){5-6} \cmidrule(lr){7-8}
& & R@1 & R@5 & R@1 & R@5 & R@1 & R@5 \\
\midrule
\multirow{4}{*}{Supervised} 
 & Sample4Geo~\cite{deuser2023sample4geo} & 80.57 & 96.53 & \textbf{86.87} & 98.71 & \textbf{91.38} & \textbf{99.74} \\
 & CAMP~\cite{wu2024camp} & 76.45 & 96.01 & 84.81 & \textbf{99.23} & 89.19 & 99.23 \\
 & DAC~\cite{xia2024enhancing} & \textbf{80.92} & \textbf{97.04} & 85.32 & 98.71 & 85.46 & 99.10 \\
 & MEAN~\cite{chen2025mean} & 78.76 & \textbf{97.04} & 85.33 & 99.10 & 89.32 & 98.97 \\
\midrule
\multirow{4}{*}{Self-supervised} & EM-CVGL~\cite{li2024learning} & 18.02 & 50.36 & 17.76 & 50.19 & 17.50 & 49.98 \\
 & CDIKTNet~\cite{chen2025limited} & 17.25 & 40.15 & 23.04 & 48.39 & 25.48 & 52.38 \\
 & DMNIL~\cite{chen2026without} & 73.10 & 92.92 & 74.65 & 95.37 & 75.03 & 95.50 \\
 & STEAM & \textbf{90.73} & \textbf{98.71} & \textbf{96.91} & \textbf{99.36} & \textbf{96.78} & \textbf{99.36} \\
\bottomrule
\end{tabularx}
}
\end{table*}

\textbf{DenseUAV.} We treat the 80, 90, and 100 m data as separate subsets for training and evaluation. In Table~\ref{tab:sota_denseuav}, STEAM achieves D2S R@1 scores of 90.73\%, 96.91\%, and 96.78\%, respectively. These exceed DMNIL by 17.63, 22.26, and 21.75 percentage points under the reported configurations. The corresponding R@5 scores are 98.71\%, 99.36\%, and 99.36\%. The difference between R@5 and R@1 is largest at 80 m, reaching 7.98 percentage points. Thus, a portion of the remaining errors at this height still place the correct location among the first five candidates. These results extend the evaluation to densely sampled low-altitude imagery and motivate examining the ordering of visually similar candidates in Section~\ref{sec:qualitative_retrieval}.

\begin{table*}[width=\FullWidth,pos=t]
\centering
\caption{Ablation study on the University-1652 dataset.
Added modules are marked with $\checkmark$.
The same dynamic threshold filtering strategy is used
in all settings.}
\label{tab:ablation}
{\small
\begin{tabularx}{\textwidth}{c|c c c c c c c|>{\centering\arraybackslash}X>{\centering\arraybackslash}X|>{\centering\arraybackslash}X>{\centering\arraybackslash}X}
\toprule
\multirow{2}{*}{Setting}
& \multicolumn{7}{c|}{Components}
& \multicolumn{2}{c|}{Drone $\rightarrow$ Satellite}
& \multicolumn{2}{c}{Satellite $\rightarrow$ Drone} \\
\cmidrule(lr){2-8} \cmidrule(lr){9-10} \cmidrule(lr){11-12}
& Backbone & SAFA & SSA & SSA$^{*}$ & BTSM & AdPu-U & AdPu-R & R@1 & AP & R@1 & AP \\
\midrule
I   & $\checkmark$ &&&&&&& 45.29 & 50.00 & 72.04 & 42.60 \\
II  & $\checkmark$ & $\checkmark$ &&&&&& 56.46 & 61.35 & 78.60 & 57.90 \\
III & $\checkmark$ && $\checkmark$ &&&&& 74.93 & 78.38 & 87.73 & 74.06 \\
IV  & $\checkmark$ &&& $\checkmark$ &&&& 79.57 & 82.61 & 92.30 & 79.51 \\
V   & $\checkmark$ &&& $\checkmark$ & $\checkmark$ &&& 90.55 & 92.07 & 92.87 & 88.72 \\
VI  & $\checkmark$ &&& $\checkmark$ & $\checkmark$ & $\checkmark$ && 93.59 & 94.61 & 94.86 & 92.28 \\
VII & $\checkmark$ &&& $\checkmark$ & $\checkmark$ & $\checkmark$ & $\checkmark$ & \textbf{95.26} & \textbf{96.08} & \textbf{96.01} & \textbf{94.21} \\
\bottomrule
\end{tabularx}
}
\end{table*}

\subsection{Ablation Studies}

\begin{figure}[pos=tbp]
  \centering
  \includegraphics[width=\columnwidth]{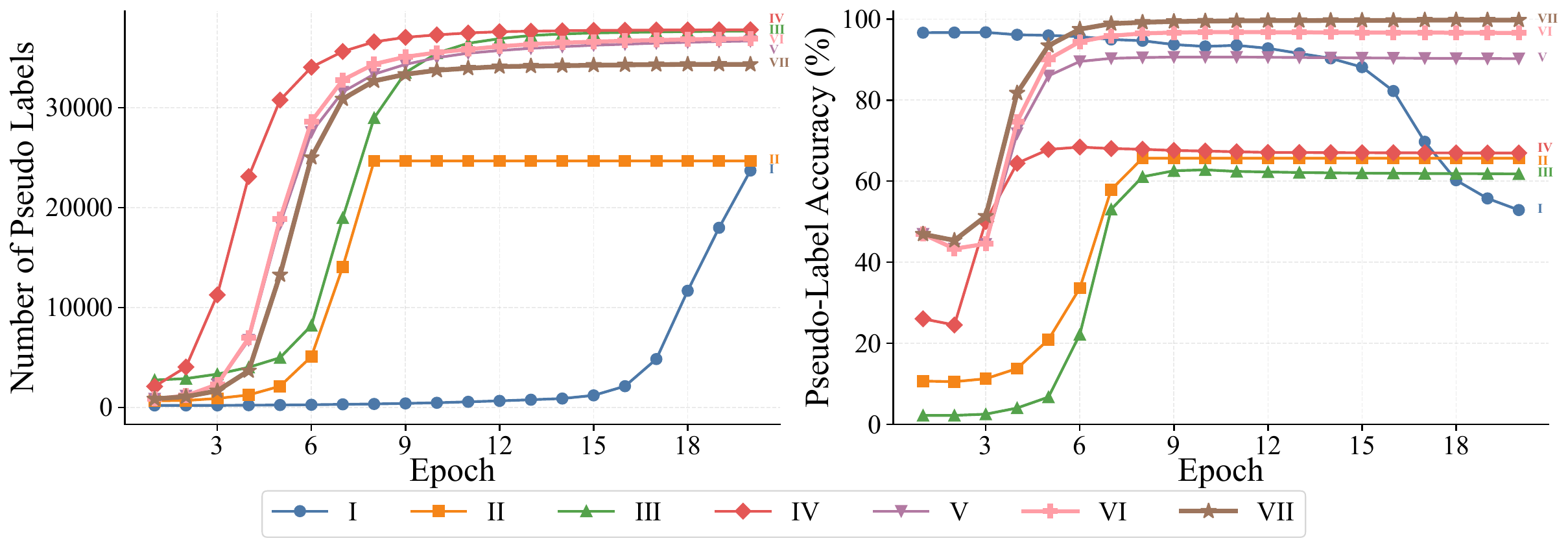}
  \caption{Number of pseudo-labels and accuracy over the complete 20-epoch training schedule under different ablation settings (I--VII correspond to Table~\ref{tab:ablation}).}
  \label{fig:ablation_pseudo}
\end{figure}

We conduct progressive ablation studies on the University-1652 dataset to evaluate the contribution of each component in STEAM. Table~\ref{tab:ablation} reports the retrieval performance, while Fig.~\ref{fig:ablation_pseudo} illustrates the evolution of pseudo-label quantity and accuracy during training. SAFA denotes the spatial-aware feature aggregation module in~\cite{li2024unleashing}; SSA is the proposed Stable Spatial-Aware Module, where SSA$^*$ indicates parameter sharing between the two branches; BTSM denotes Bidirectional Top-$K$ Soft Matching; and AdPu-U and AdPu-R denote Confidence-aware Update and Expired Label Removal, respectively. For pseudo-label accuracy analysis, predicted correspondences are compared with ground-truth correspondences. Ground-truth correspondences are not inputs to pseudo-label mining, mini-batch construction, or gradient-based optimization.

Replacing SAFA with SSA (II$\rightarrow$III) improves D2S R@1
from 56.46\% to 74.93\%.
As shown in Fig.~\ref{fig:ablation_pseudo}, SSA produces more
pseudo-labels, indicating increased pseudo-label coverage.
Enabling parameter sharing (III$\rightarrow$IV) further increases
D2S R@1 from 74.93\% to 79.57\%.

Introducing BTSM (IV$\rightarrow$V) yields a substantial performance improvement, boosting D2S R@1 to 90.55\%. Meanwhile, the pseudo-label accuracy increases significantly to around 90\%, suggesting that elastic matching effectively expands pseudo-label coverage without introducing excessive noise.

Applying AdPu-U (V$\rightarrow$VI) further improves pseudo-label quality by replacing historical pseudo-labels with higher-confidence matches. Consequently, the pseudo-label accuracy increases from approximately 90\% to over 96\%, which is accompanied by a further improvement of D2S R@1 to 93.59\%. Finally, AdPu-R (VI$\rightarrow$VII) removes outdated pseudo-labels that are no longer supported by the current feature space, maintaining the pseudo-label accuracy above 99\% during the later training stage and achieving the best retrieval performance among the evaluated ablation settings, with 95.26\% D2S R@1 and 96.01\% S2D R@1.

The progression shows that descriptor improvements leave further gains available from correspondence selection and maintenance. After SSA and parameter sharing, BTSM contributes another 10.98 percentage points in D2S R@1, and the two AdPu steps together contribute 4.71 points. These gains accompany the reported changes in pseudo-label coverage and accuracy. The results support complementary roles for representation learning, candidate discovery, and repository maintenance within the evaluated pipeline.

\subsection{Influence of $K$ in Elastic Matching}
\begin{figure}[pos=tbp]
  \centering
  \includegraphics[width=\columnwidth]{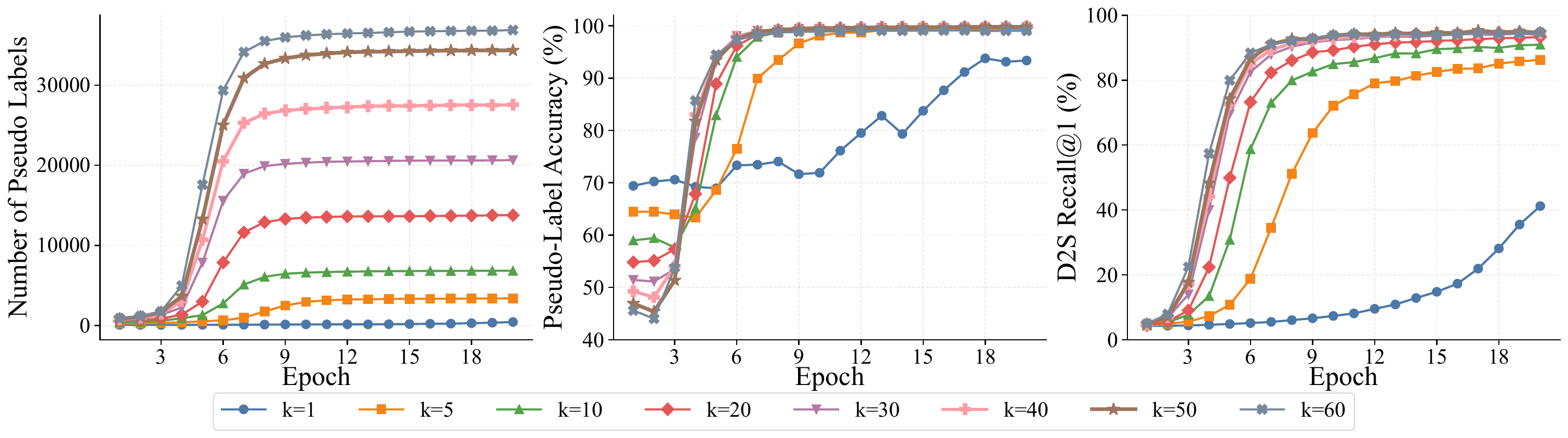}
  \caption{Number of pseudo-labels, pseudo-label accuracy and validation D2S Recall@1 over the complete 20-epoch training schedule under different $K$ settings.}
  \label{k_analysis}
\end{figure}

We evaluate $K \in \{1,5,10,20,30,40,50,60\}$. Fig.~\ref{k_analysis} reports pseudo-label quantity and accuracy
over the combined training and validation sets, together with
validation D2S Recall@1. Validation D2S Recall@1 guides the selection of $K$ and checkpoints. For the selected $K=50$, the pseudo-label quantity and accuracy also indicate a favourable balance between coverage and precision.

\begin{figure}[pos=tbp]
  \centering
  \includegraphics[width=\columnwidth]{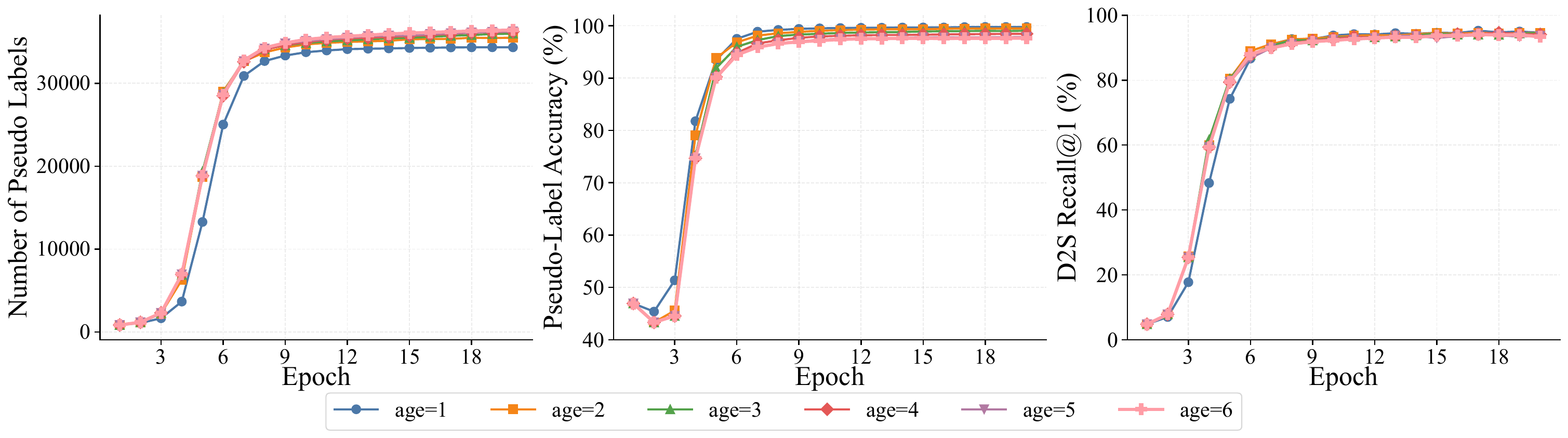}
  \caption{Number of pseudo-labels, pseudo-label accuracy and validation D2S Recall@1 over the complete 20-epoch training schedule under different $M_e$ settings.}
  \label{fig:age_analysis}
\end{figure}

\begin{figure*}[width=\textwidth,pos=t]
\centering
\includegraphics[width=\textwidth]{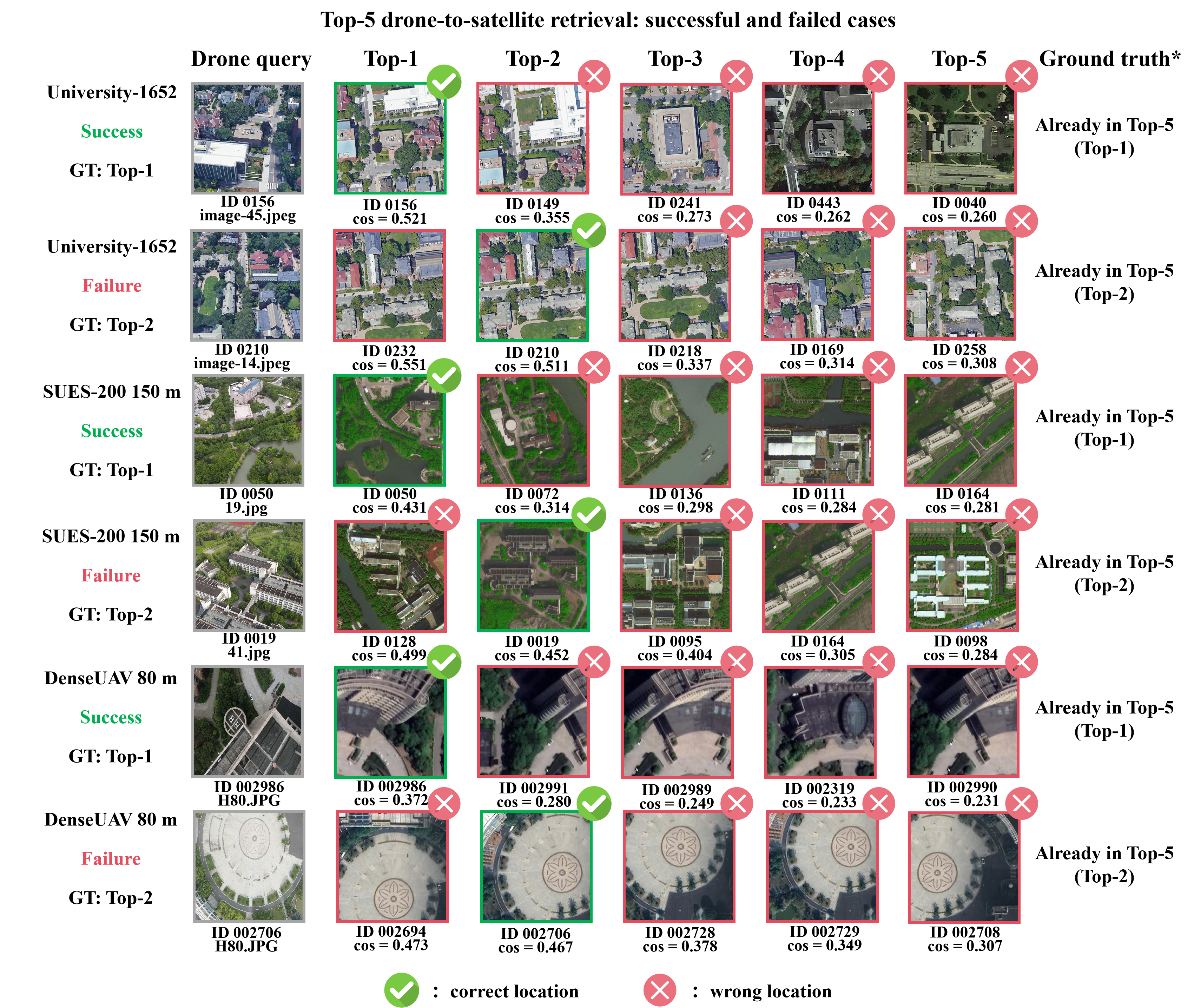}
\caption{Drone-to-Satellite Top-5 retrieval results on University-1652, SUES-200 (150 m), and DenseUAV (80 m). Each dataset includes a correct Top-1 example and a Top-1 error with the ground truth at Top-2. Each row shows a drone query followed by satellite candidates in descending order of cosine similarity. Green borders and checkmarks indicate correct matches; red borders and crosses indicate incorrect matches. Candidate location IDs and cosine similarities are shown below the images. GT denotes the rank of the ground-truth match.}
\label{fig:top5_retrieval}
\end{figure*}

When $K=1$, matching degenerates to strict mutual nearest-neighbor selection, resulting in fewer than 500 pseudo-labels even after 20 epochs, although the accuracy remains above 90\%. As $K$ increases, more candidate correspondences are included, leading to a substantial increase in pseudo-label quantity. At $K=50$, the number of pseudo-labels reaches approximately 34,000 while the final accuracy reaches 99.78\%, indicating that elastic matching expands coverage without sacrificing pseudo-label quality. Increasing $K$ to 60 provides only a modest increase in pseudo-label quantity while slightly degrading both pseudo-label accuracy and retrieval performance. These results suggest a clear coverage-noise trade-off: increasing $K$ expands pseudo-label coverage, beyond which ambiguous correspondences degrade retrieval performance.

\subsection{Influence of $M_e$ in Adaptive Purification}
We evaluate maximum pseudo-label age $M_e \in \{1,2,3,4,5,6\}$. Fig.~\ref{fig:age_analysis} reports pseudo-label quantity and accuracy over the combined training and validation sets, together with validation D2S Recall@1. We select $M_e=1$ using validation D2S Recall@1. Its pseudo-label quantity and accuracy also indicate a favourable balance between repository coverage and pseudo-label quality. With $M_e=1$, an existing entry expires after one epoch if its stored pseudo-label is neither supported by an agreeing candidate
nor successfully replaced. This strategy maintains the highest pseudo-label accuracy (around 99.8\%) while keeping the repository size stable at approximately 34,300 pseudo-labels. Increasing $M_e$ allows stale pseudo-labels to persist longer, leading to a gradual increase in repository size but a slight decrease in pseudo-label accuracy. For $M_e \ge 3$, the repository continues to grow modestly, while the final pseudo-label accuracy gradually declines. Overall, the performance is relatively insensitive to $M_e$.

The role of $M_e$ is therefore to control how long unsupported assignments remain eligible for optimization. A larger repository is useful only when its additional labels remain reliable. The selected value reflects this balance on University-1652 under the evaluated training schedule.

\subsection{Reproducibility across Random Seeds}
\label{sec:seed_reproducibility}
To assess variation across random seeds, we report three STEAM runs
on University-1652 using seeds 42, 43, and 2026.
Table~\ref{tab:random_seeds} reports the individual results and their
mean and sample standard deviation.

\begin{table}[pos=tbp]
\centering
\caption{Retrieval performance across random seeds on University-1652 (\%). The last row reports the mean $\pm$ sample standard deviation over three runs.}
\label{tab:random_seeds}
{\small
\setlength{\tabcolsep}{1pt}
\begin{tabularx}{\columnwidth}{c *{4}{>{\centering\arraybackslash}X}}
\toprule
\multirow{2}{*}{Seed} & \multicolumn{2}{c}{Drone $\rightarrow$ Satellite} & \multicolumn{2}{c}{Satellite $\rightarrow$ Drone} \\
\cmidrule(lr){2-3} \cmidrule(lr){4-5}
& R@1 & AP & R@1 & AP \\
\midrule
42 & 95.26 & 96.08 & 96.01 & 94.21 \\
43 & 95.12 & 95.91 & 95.92 & 94.63 \\
2026 & 95.01 & 95.85 & 95.58 & 94.80 \\
\midrule
Mean$\pm$SD & 95.13$\pm$0.13 & 95.95$\pm$0.12 & 95.84$\pm$0.23 & 94.55$\pm$0.30 \\
\bottomrule
\end{tabularx}
}
\end{table}

Across these runs, the mean D2S R@1 and AP are
$95.13\pm0.13$\% and $95.95\pm0.12$\%, respectively.
The corresponding S2D results are $95.84\pm0.23$\% and
$94.55\pm0.30$\%.
The largest observed R@1 differences are 0.25 and 0.43 percentage
points for D2S and S2D, respectively.
These results indicate limited variation across the three reported runs.

\subsection{Qualitative Retrieval Analysis}
\label{sec:qualitative_retrieval}

Figure~\ref{fig:top5_retrieval} illustrates successful retrieval and
Top-1 errors for the corresponding models on the three dataset settings.
In the successful University-1652 example, the first two satellite
candidates show substantial scene overlap, yet the correct location
is ranked first.
Its Top-1 error involves similar building groups and overlapping
campus layouts.
The SUES-200 and DenseUAV errors involve similar building groups and
repeated circular pavement patterns, respectively, with the correct
matches ranked second.
In each illustrated pair of cases, the Top-1 error has a smaller
top-two similarity margin than the successful query.

We define the top-two retrieval margin as $\Delta=s_1-s_2$, where
$s_1$ and $s_2$ are the largest and second-largest cosine similarities.
Across the full test sets of University-1652, SUES-200 (150 m), and
DenseUAV (80 m), mean margins for correct Top-1 queries are
0.163, 0.124, and 0.095, respectively.
The corresponding averages for Top-1 errors are 0.022, 0.020, and 0.023.
Among Top-1 errors, 51.4\%, 58.6\%, and 65.3\% are recovered at
Top-2 in the three dataset settings, respectively.
These observations indicate that smaller similarity margins are
associated with retrieval errors, while ranking visually similar
candidates remains challenging.

\subsection{Limitations}

Visually similar locations, including repeated building groups, repetitive textures,
and overlapping scene layouts, can still produce incorrect Top-1 rankings
(Fig.~\ref{fig:top5_retrieval}).
In addition, STEAM extracts full-dataset features and computes cross-view
similarities at the beginning of every training epoch.
This repeated mining step incurs computational and memory costs that increase
with dataset size. More efficient pseudo-label mining remains a direction for
future work.

\section{Conclusion}
In this paper, we propose STEAM, a unified self-training framework for UCVGL. STEAM combines end-to-end representation learning with iterative pseudo-label refinement, without image-generation-based warm-up or clustering.
By integrating a Stable Spatial-Aware Module, Elastic Matching, and Adaptive Purification, STEAM progressively builds reliable cross-view correspondences while effectively mitigating the accumulation and propagation of noisy pseudo-labels.
STEAM achieves the highest R@1 and AP among the compared unsupervised methods on University-1652 and competitive retrieval performance on SUES-200 and DenseUAV.

\section*{Funding}
This research did not receive any specific grant from funding agencies in the public, commercial, or not-for-profit sectors.

\section*{Declaration of competing interest}
The authors declare that they have no known competing financial interests or personal relationships that could have appeared to influence the work reported in this paper.

\section*{Data availability}
The datasets used in this study are available from their original providers: University-1652~\cite{zheng2020university} at \url{https://github.com/layumi/University1652-Baseline}, SUES-200~\cite{zhu2023sues} at \url{https://github.com/Reza-Zhu/SUES-200-Benchmark}, and DenseUAV~\cite{dai2024vision} at \url{https://github.com/Dmmm1997/DenseUAV}.

\bibliographystyle{elsarticle-num}
\bibliography{cas-refs}
\end{document}